\documentclass[10pt,twocolumn,letterpaper]{article}

\usepackage{cvpr}              
\usepackage{graphicx}
\usepackage{amsmath}
\usepackage{amssymb}
\usepackage{booktabs}

\usepackage{float}
\usepackage{multirow}
\usepackage{placeins}
\usepackage{color, colortbl}
\usepackage{xcolor}
\definecolor{dgreen}{RGB}{1,150,74}
\newcommand\best[1]{\textcolor{red}{\textbf{#1}}}
\newcommand\bb[1]{\textbf{#1}}
\newcommand\second[1]{\textcolor{cyan}{\textbf{#1}}}

\newcommand\blue[1]{\textcolor{blue}{#1}}

\newcommand\up[1]{\textcolor{dgreen}{$\uparrow{#1}$}}

\newcommand\down[1]{\textcolor{red}{$\downarrow{#1}$}}
\newcommand\mypar[1]{\par\vspace{1.5mm}\noindent\textbf{#1}\;\;}

\usepackage[pagebackref,breaklinks,colorlinks]{hyperref}

\usepackage[capitalize]{cleveref}
\crefname{section}{Sec.}{Secs.}
\Crefname{section}{Section}{Sections}
\Crefname{table}{Table}{Tables}
\crefname{table}{Tab.}{Tabs.}

\def\cvprPaperID{3593} 
\def\confName{CVPR}
\def\confYear{2022}

\begin{document}

\title{Embracing Single Stride 3D Object Detector with Sparse Transformer}

\author{
Lue Fan\\
CASIA\\
{\tt\small fanlue2019@ia.ac.cn}
\and
Ziqi Pang\\
UIUC\\
{\tt\small ziqip2@illinois.edu}
\and
Tianyuan Zhang\\
CMU\\
{\tt\small tianyuaz@andrew.cmu.edu}
\and
Yu-Xiong Wang\\
UIUC\\
{\tt\small yxw@illinois.edu}
\and
Hang Zhao\\
THU\\
{\tt\small hangzhao@mail.tsinghua.edu.cn}
\and
Feng Wang\\
TuSimple\\
{\tt\small feng.wff@gmail.com}
\and
Naiyan Wang\\
TuSimple\\
{\tt\small winsty@gmail.com}
\and
Zhaoxiang Zhang\\
CASIA\\
{\tt\small zhaoxiang.zhang@ia.ac.cn}
}
\maketitle

\begin{abstract}
   In LiDAR-based 3D object detection for autonomous driving, the ratio of the object size to input scene size is significantly smaller compared to 2D detection cases.
Overlooking this difference, many 3D detectors directly follow the common practice of 2D detectors, which downsample the feature maps even after quantizing the point clouds.
In this paper, we start by rethinking how such multi-stride stereotype affects the LiDAR-based 3D object detectors. Our experiments point out that the downsampling operations bring few advantages, and lead to inevitable information loss. To remedy this issue, we propose Single-stride Sparse Transformer (SST) to maintain the original resolution from the beginning to the end of the network. Armed with transformers, our method addresses the problem of insufficient receptive field in single-stride architectures. It also cooperates well with the sparsity of point clouds and naturally avoids expensive computation. Eventually, our SST achieves state-of-the-art results on the large-scale Waymo Open Dataset. It is worth mentioning that our method can achieve exciting performance (\textbf{83.8} LEVEL\_1 AP on validation split) on small object (pedestrian) detection due to the characteristic of single stride. Codes will be released at \url{https://github.com/TuSimple/SST}.

\end{abstract}

\section{Introduction}
\begin{figure}[t]
    \vspace{5mm}
    \centering
    \includegraphics[width=\columnwidth]{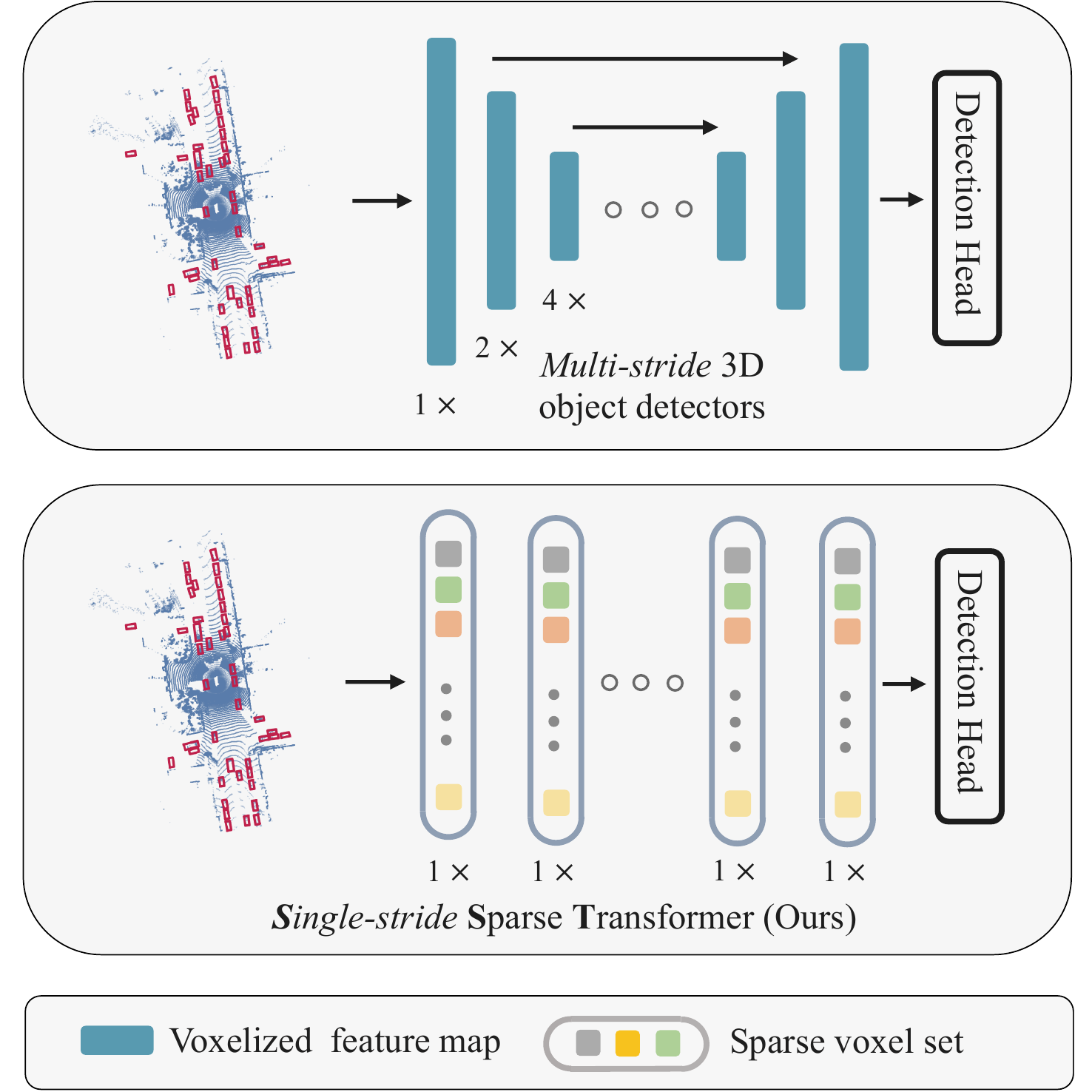}
    \caption{Compared with previous multi-stride 3D detectors, our model is single-stride and operates sparsely on the non-empty voxels. We paint the vehicle bounding boxes on the input point cloud to show the tiny object size compared to the  input scene size.}
    \vspace{-4mm}
    \label{fig:fig1}
\end{figure}

\begin{figure}[t]
    \centering
    \includegraphics[width=0.99\columnwidth]{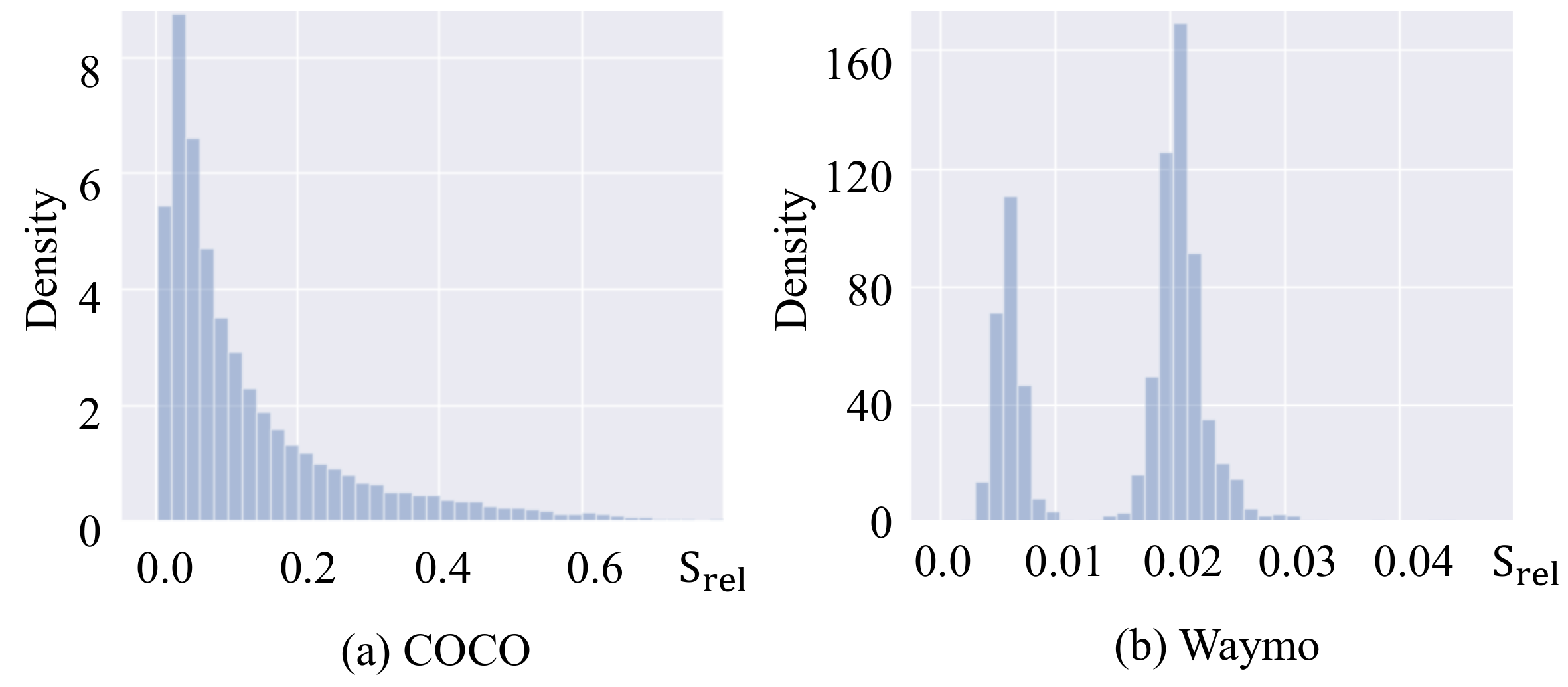}
    \caption{\textbf{Distribution of the relative object size $S_{rel}$ in COCO dataset~\cite{coco} and Waymo Open Dataset (WOD).} $S_{rel}$ is defined as $\sqrt{A_{o}/A_{s}}$, where $A_o$ denotes the area of 2D objects (COCO) and the BEV area of 3D objects (WOD). $A_s$ is the image area in COCO, and $150m \times 150m$ in WOD. In COCO 73.03\% objects in COCO have a $S_{rel}$ larger than 0.04, while only 0.54\% objects in WOD have a $S_{rel}$ larger than 0.04.}
    \label{fig:hist}
\vspace{-5mm}
\end{figure}

\label{sec:intro}
LiDAR-based 3D object detection for autonomous driving has been benefiting from the progress of image-based object detection. The mainstream 3D detectors quantize the 3D space into a stack of pseudo-images from Bird Eye's View (BEV), which makes it convenient to borrow advanced techniques from the 2D counterparts. Many works~\cite{pointpillar, second, centerpoint, rangedet} are proposed under this paradigm and achieve competitive performance.
However, 3D and 2D spaces have intrinsic distinction in their relative object scales, where the objects in 3D spaces have much smaller relative sizes (See Fig.~\ref{fig:hist}). For example, in Waymo Open Dataset~\cite{wod}, the perception range is usually $150m \times 150m$, while a vehicle is only about $4m$ long, even a pedestrian occupies as little as $1m$ in length. Such a tiny pedestrian equivalently translates to an object of size $8 \times 8$ pixels in a $1200 \times 1200$ image, suggesting that object detection on such a tiny scale is one of the challenges in 3D object detection. 

\par
Different from the above challenge of small scales in the 3D space, 2D detectors have to consider the handling of the objects with varied scales.
It is observed in Fig.~\ref{fig:hist} that the scales of objects in 2D images exhibit a long-tail distribution, while in 3D space they are quite concentrated due to the non-projective transformation used in voxelization.
To handle the varied scales, 2D detectors~\cite{fpn, snip, sniper, tridentnet} usually build multi-scale features with a series of downsampling and upsampling operations.
Such multi-scale architecture is also widely inherited in 3D detectors (See Fig.~\ref{fig:fig1})~\cite{pointpillar, second, voxelnet, centerpoint, rangedet}.
Since the object size in 3D object detectors is usually tiny while no large objects exist, a question naturally arises: \textit{do we really need downsampling in 3D object detectors ?}
\par
With this question in mind, we make an exploratory attempt on the single-stride architecture with \textbf{no downsampling operators}. The single-stride network maintains the original resolution throughout the network. However, it is challenging to make such a design feasible.
The discard of downsampling operators leads to two issues: 1) the increase of computation cost; 2) the decrease of receptive field. 
The former constrains the applicability to the real-time system and the latter hinders the capability of object recognition.
For the issue of computation, sparse convolution seems to be a solution, but the sparse connectivity between voxels\footnote{We provide a clear illustration for this in our supplementary materials.} makes the decrease of receptive field even more severe (See Table~\ref{tab:comp_with_ss_pp}).
For the issue of receptive field, we experimentally show that some commonly adopted techniques do not meet our needs (See Table~\ref{tab:pilot}):
the dilated convolution~\cite{deeplab,dilation} is not friendly to small objects, and the larger kernel leads to unaffordable computational overhead in the single stride architecture.
Therefore, we are getting into a dilemma, where it is difficult to design a convolutional network simultaneously satisfying the three aspects: \textit{single stride architecture}, \textit{sufficient receptive field}, and \textit{acceptable computation cost}.

\par
These difficulties naturally lead us to think out of the paradigm of CNN, and the attention mechanism emerges as a better option because of the following two reasons: 
1) The attention-based model is better at capturing large context and build sufficient receptive field.
2) Due to the capability of modeling dynamic data, the attention-based model fits well into the sparse voxelized representation of point clouds, where only a small portion of voxels are occupied. This property guarantees the efficiency of our single stride network.
Although the attention mechanism is efficient on sparse data, computing attentions on a global scale is still unaffordable and undesirable. So we partition the voxelized 3D space into many local regions and apply self-attention inside each of them. Eventually, this local attention mechanism, named as \textbf{Sparse Regional Attention (SRA)}, enjoys the best of two worlds. By stacking SRA layers, we make the single-stride network feasible and obtain a transformer-style network, called \textbf{Single-stride Sparse Transformer (SST)}. 
Extensive experiments are conducted on the large-scale Waymo Open Dataset~\cite{wod}.
We summarize our contributions as follows:
\begin{itemize}
\item 
We rethink the architecture of current mainstream LiDAR-based 3D detectors. With pilot experiments, we point out that the network stride is an overlooked design factor for LiDAR-based 3D detectors.  
\item 
We propose the {\bf S}ingle-stride {\bf S}parse {\bf T}ransformer (\textbf{SST}). With its local attention mechanism and capability of handling sparse data, we overcome receptive field shrinkage in the single-stride setting and avoid heavy computational overhead.
\item
Our method achieves state-of-the-art performance on the large-scale Waymo Open Dataset. Thanks to the characteristic of single stride, our method obtains exciting results on tiny objects like pedestrians (83.8 LEVEL\_1 AP on the validation split).
\end{itemize}

\section{Related Work}
\mypar{3D LiDAR-based Detection}
There are three major representations for point cloud learning in autonomous driving, Point-based, Voxel-based, and Range View. 
Point based representation backed by PointNet families \cite{pointnet, pointnet++} are widely adopted for feature learning of small region of irregular points \cite{frustumpointnet,lidarrcnn,pointrcnn,fastpointrcnn}. 
Voxel-based representation \cite{voxelnet, centerpoint, second, pointpillar} combined with convolutions are the most popular treatment.  
As explored in several recent works \cite{velofcn,  lasernet, rangedet,tothepoint, rcd}, range view enjoys computational advantages over voxels, especially for long-range LiDAR sensors. 
Some hybrid approaches investigate how to combine different types of representations \cite{pvrcnn, pvrcnnpp, rsn, lidarrcnn,mv3d, pillarbased}.

\mypar{Transformers in Visual Recognition}
The success of transformer architectures in NLP \cite{transformer, bert} and speech recognition\cite{attentionspeech} has inspired lots of work to investigate the power of attention in visual recognition \cite{standalone, explore_self_att, vit, swin, deit}. 
The pioneering work ViT \cite{vit} splits an image into patches, and then feeds sequences of patches to multiple transformer blocks for image classification. DeiT \cite{deit} explores training strategies for data-efficient learning of vision transformers. Swin-Transformer~\cite{swin}  exploits the power of local attention to build high-performance transformer-based image backbones. 
Several works have investigated the use of transformers for point cloud perceptions. Some of them focus on the indoor scene such as~\cite{pointtransformer, pct, 3detr}. For autonomous driving scenarios, Pointformer \cite{pointformer} proposes a point-based local and global attention module directly operating on point clouds. In addition, VoTr \cite{voxelformer} uses the local self-attention module to replace the sparse convolution~\cite{sparseconv} for voxel processing, where each voxel serves as a query and attends with its neighbor voxels. 

\mypar{Small Object Detection} Small object detection~\cite{dota, visdrone, coco} is a challenging track in 2D object detection. The mainstream of current methods~\cite{hrnet, querydet, dssd, fpn, beyondskip} focuses on increasing the resolution of the input and output features, while none of them gives up the multi-stride architectures. Some other methods adopt the scale-aware training~\cite{fpn, sniper, tridentnet} and strong data augmentations~\cite{augforsmall, learnaug}. To the best of our knowledge, there is no method specialized for small object detection in 3D space.
\section{Discussion of Network Stride}
\label{sec:pilot}
The stride of a network is a simple but critical aspect in the architecture design. 
Some previous works~\cite{centerpoint, rangedet, afdet} in 3D detection have found that the performance can benefit from the recovery of output resolution by upsampling. 
However, they do not delve into this phenomenon.
Therefore, we conduct a simple pilot study to reveal the influence of network stride on 3D detectors and motivate the design of our network.
\par
For generality, we adopt the widely used PointPillars~\cite{pointpillar} in MMDetection3D~\cite{mmdet3d} as our base model.
The experiments are conducted on Waymo Open Dataset~\cite{wod}.
We uniformly sample 20\% training data (32K frames) \footnote{Training with 20\% data is a setting for efficient validation adopted in~\cite{openpcdet2020,mmdet3d}.} and adopt $1\times$ schedule (12 epochs).
\par
Based on the standard PointPillars model $D_2$, we extend it to three more variants: $D_3$, $D_1$, and $D_0$, and they only differ in the network stride.
From $D_3$ to $D_0$, the set of strides of their four stages for each model are $\{1, 2, 4, 8\}$, $\{1, 2, 4, 4\}$, $\{1, 2, 2, 2\}$ and $\{1, 1, 1, 1\}$, respectively.
Since the output feature maps of the four stages will be upsampled to the original resolution by an FPN-like module, our modification does not change the resolution of feature maps in the detection head. Except for the resolution of feature maps, all the four models have the same hyper-parameters. To reduce memory overhead, we change the filter number from 256 to 128 in convolution layers.
\par
The main results are shown in Table~\ref{tab:pilot}. Performances of all three classes improve from $D_3$ to $D_1$, and there is a significant boost from $D_2$ to $D_1$. 
The performance boost from $D_3$ to $D_1$ supports our motivation that \emph{Smaller strides are better for 3D detection.} 
\par
However, from $D_1$ to $D_0$, the vehicle performance has a significant drop, while the performance drop in pedestrian is slight and performance of cyclist keeps going up. 
We conjecture that the limited receptive field of $D_0$ hinders the performance improvement from $D_1$ to $D_0$ since the pedestrian and cyclist have smaller sizes than vehicles.
\par
To verify our conjecture, we add two more variants: $D_0^{dilation}$ and $D_0^{5\times5}$. 
$D_0^{dilation}$ adopts dilated convolutions with dilation as 2 in the last two stages. 
$D_0^{5\times5}$ increases the kernel size in last two stages to $5 \times 5$. 
Table~\ref{tab:pilot} shows that, dilation increases the performance of vehicle class while decreases performances of pedestrian and cyclist, indicating that it indeed enlarges the receptive field, however misses fine-grained details. 
Meanwhile, larger kernel consistently improves the performance of all three classes but unfortunately has the highest latency. 
Above studies support our major motivation of single-stride 3D detectors, and it also reveals the another important aspect in our network design: \emph{Sufficient receptive field is crucial.}
\par
\begin{table}[H]
\small
\begin{center}
\resizebox{0.9\linewidth}{!}{
\begin{tabular}{l@{\ }c@{ }c@{ }c@{\ }c@{\ }}
  \specialrule{0.7pt}{0pt}{1pt}
  \specialrule{0.7pt}{0pt}{2pt}
  Models & Vehicle & Pedestrian & Cyclist & Latency\\
    \specialrule{0.5pt}{1pt}{2pt}
$D_3$ & 63.66 & 60.82 & 47.08 & 58ms\\
    \specialrule{0pt}{1pt}{1pt}
$D_2$ & 64.01 \up{0.35} & 60.85 \up{0.03} & 47.52 \up{0.44} & 60ms\\
    \specialrule{0pt}{1pt}{1pt}
$D_1$ & 66.03 \up{2.02} & 65.06 \up{4.21} & 52.97 \up{5.45} & 91ms\\
    \specialrule{0pt}{1pt}{1pt}
$D_0$ & 64.69 \down{1.34} & 64.32 \down{0.74} & 53.02 \up{0.05} & 185ms\\
    \specialrule{1pt}{3pt}{3pt}
$D_0^{dilation}$ & 66.26 \up{1.57} & 63.51 \down{0.81} & 50.95 \down{2.07} & 192ms\\
$D_0^{5\times5}$ & 66.42 \up{1.77} & 65.71 \up{1.41} & 53.70 \up{0.68} & 340ms\\
  \specialrule{0.7pt}{2pt}{1pt}
  \specialrule{0.7pt}{0pt}{1pt}
 
 \end{tabular}
}
\end{center}
\caption{Results of pilot study on Waymo Open Dataset validation split. Latency is evaluated in 2080Ti GPU with 2000 samples after a cold start of 500 samples. For $D_n$, the arrows indicate the performance changes based on $D_{n+1}$. For $D_0^{dilation}$ and $D_0^{5 \times 5}$, the arrows indicate performance changes based on $D_0$. Best viewed in color. }
\label{tab:pilot}
\end{table}
In summary, above experiments verify two motivations of 3D object detector designs:
\begin{itemize}
    \item The single stride architecture has a great potential in LiDAR-based 3D detection for autonomous driving.
    \item The key to make single stride architecture feasible lies in appropriately addressing the shrinkage of receptive field and reducing computational overhead.
\end{itemize}

\section{Methodology}

\begin{figure*}
	\centering
	\includegraphics[width=1.8\columnwidth]{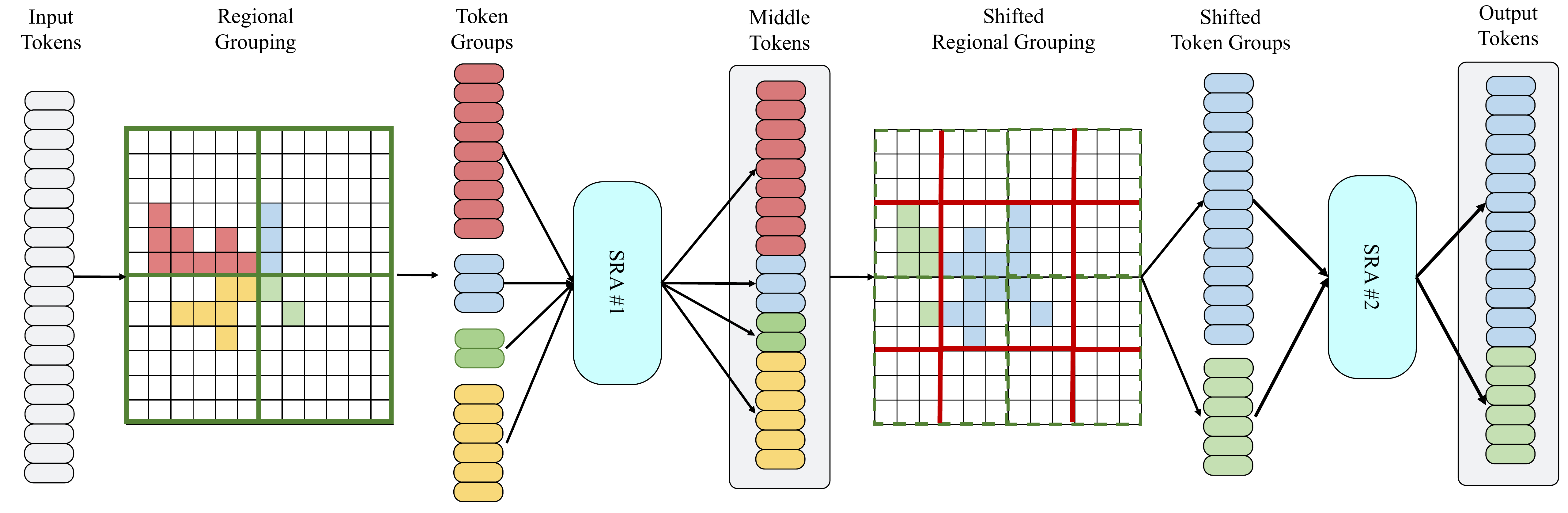}
	\caption{Computation of an example block in SST. For an incoming set of tokens, \emph{Regional Grouping} first groups them according to the partitions of regions (in Sec.~\ref{sec:grouping}). Second, \emph{Sparse Regional Attention} (SRA) deals with each group of tokens separately (Sec.~\ref{sec:method_ssr}). Third, the tokens are grouped another time according to \emph{Region Shift}, and a second SRA processes the new groups of tokens (Sec.~\ref{sec:method_region_shift}). These three steps complete the computation of a block.}
	\label{fig:ssr}
	\vspace{-8px}
\end{figure*}
\begin{figure}
	\centering
	\includegraphics[width=1.0\columnwidth]{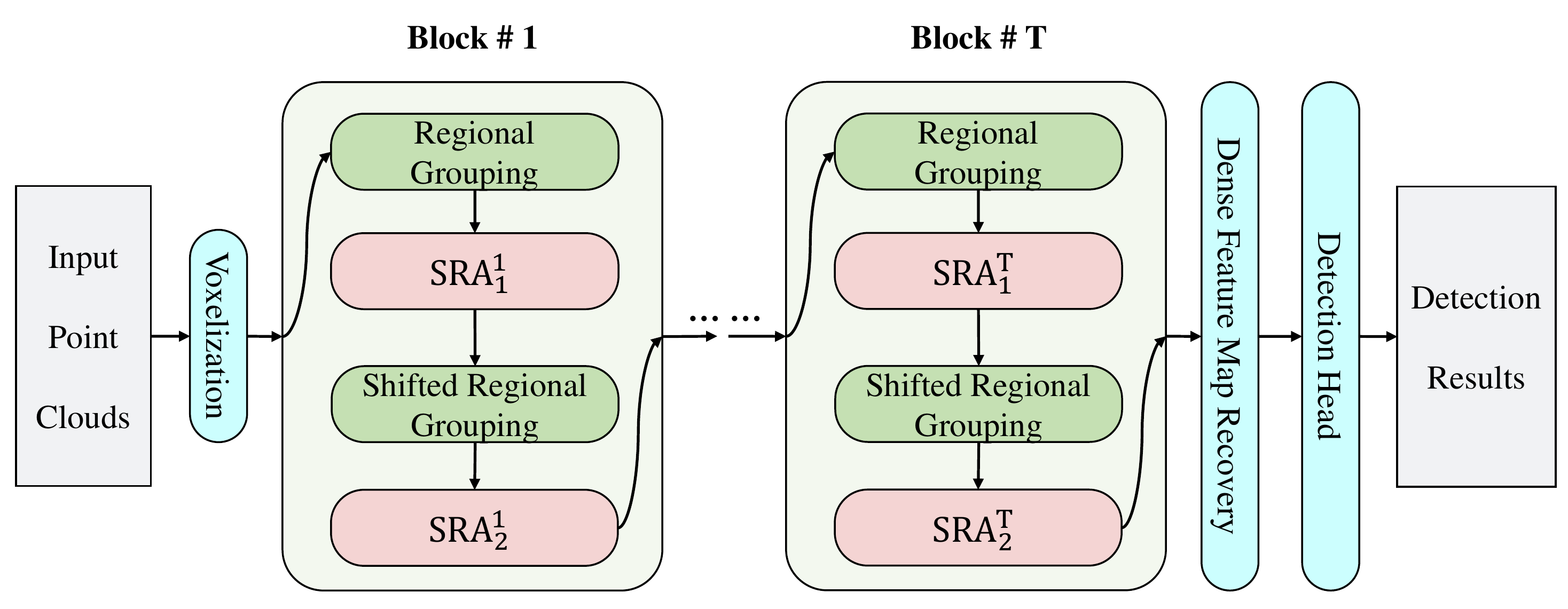}
	\caption{Architecture overview for \emph{Single-stride Sparse Transformer} (SST). It begins from voxelizing an input point cloud, then processes the voxels with $T$ blocks, and eventually recover a dense feature map. Inside each block, we consecutively append regional grouping on the voxel tokens and \emph{Sparse Regional Attention} (SRA) to process them. Details in Sec.~\ref{sec:method_arch}.}
	\label{fig:arch}
	\vspace{-8px}
\end{figure}

\subsection{Overall Architecture}
\label{sec:method_arch}
\par

So far, we know the keys to make single stride architecture feasible are sufficient receptive field and acceptable computational cost.
However, as we discussed in Sec.~\ref{sec:intro}, it is difficult to simultaneously satisfy the two factors with convolutional single stride architecture.
So we turn to the attention mechanism in Transformer~\cite{transformer}, and present our method as follows.

We build up our  \emph{Single-stride Sparse Transformer} (SST) as  in Fig.~\ref{fig:arch}.
SST voxelizes the point clouds and extracts voxel features following prior work~\cite{pointpillar, second, voxelnet}. 
For each voxel and its features, SST treats them as ``tokens.'' 
SST first partitions the voxelized 3D space to fixed-size non-overlapping regions  (Sec.~\ref{sec:grouping}).
Then SST applies \emph{Sparse Regional Attention} (SRA) to voxel tokens in each region (Sec.~\ref{sec:method_ssr}).
To handle the objects scattering multiple regions and capture useful local context, we adopt \emph{Region Shift} (Sec.~\ref{sec:method_region_shift}), which is inspired by the shifted window in Swin-Transformer~\cite{swin}.
The backbone preserves the number of voxels as well as their spatial locations, thus satisfying the single-stride property, and can be integrated with mainstream detection heads (Sec.~\ref{sec:method_integrate_detection}).

\subsection{Regional Grouping}
\label{sec:grouping}

Given the input voxel tokens, \emph{Regional Grouping} divides the 3D space into non-overlapping regions, 
so that the self-attentions only interact with tokens coming from the same regions.
The regional grouping not only maintains sufficient receptive field, but also avoids expensive computation overhead in global attentions. We illustrate it intuitively in Fig.~\ref{fig:ssr}. Each \emph{regional grouping} divides the input tokens into groups according to their physical locations, where the tokens belonging to the same regions (green rectangles) are assigned to the same group. 

\subsection{Sparse Regional Attention}
\label{sec:method_ssr}

\emph{Sparse Regional Attention} (SRA) operates on the regional sparse sets of voxel tokens coming from \emph{regional grouping}. 
For a group of tokens $\mathcal{F}$ and their corresponding spatial $(x, y, z)$ coordinates $\mathcal{I}$, SRA follows conventional transformers as follows
\begin{equation}
	\begin{aligned}
		\label{eq:method_sra}
		& \mathcal{F}^{'} = \mathbf{MSA}(\mathbf{LN}(\mathcal{F}), \mathbf{PE}(\mathcal{I})) + \mathcal{F} \\
		& \widetilde{\mathcal{F}} = \mathbf{MLP}(\mathbf{LN}(\mathcal{F}^{'})) + \mathcal{F}^{'} \\
	\end{aligned}
\end{equation}
where $\mathbf{PE}(\cdot)$ stands for the absolute positional encoding function used in~\cite{detr}, 
$\mathbf{MSA}(\cdot)$ denotes the Multi-head Self-Attention, 
and $ \mathbf{LN}(\cdot) $ represents Layer Normalization.   
This manner of SRA well exploits the sparsity of point clouds, because it only computes the voxels with actual LiDAR points.

\mypar{Region Batching for Efficient Implementation} Due to the sparsity of point cloud, the number of valid tokens in each region varies. To utilize the parallel computation of modern devices, we batch regions with similar number of tokens together. In practice, if a region contains the tokens with number $N_{token}$, satisfying:
\begin{equation}
    2^i \leqslant N_{token} < 2^{i+1}, \quad i \in \{0, 1, 2, 3, 4, 5, 6\},
\end{equation}
then we pad the number of tokens to $2^{i+1}$. With padded tokens, we can divide all the regions into several batches, and then process all regions in the same batch in parallel. As the padded tokens are masked in the computation as in~\cite{detr, transformer}, they have no effect on other valid tokens.
In this way, it is easy to implement an efficient SRA module in current popular deep learning frameworks without engineering efforts as taken in the sparse convolution~\cite{sparseconv, second}.
\subsection{Region Shift}
\label{sec:method_region_shift}
Though SRA can cover a considerably large region, there are some objects inevitably truncated by the grouping.
To tackle this issue and aggregate useful context, we further use \emph{Region Shift} in our design, which is similar to the shifting mechanism in Swin Transformer for information communication.
Supposing the size of regions in \emph{regional grouping} is $(l_x, l_y, l_z)$, the \emph{Region Shift} moves the original regions by $(l_x/2, l_y/2, l_z/2)$ and groups the tokens according to this new set of regions, as illustrated in ``Shifted regional grouping'' of Fig.~\ref{fig:ssr}. 
\subsection{Integration with Detection}
\label{sec:method_integrate_detection}

To work with the existing detector heads, SST places the sparse voxel tokens back to dense feature maps according to their spatial locations. Unoccupied locations are filled with zeros.
As LiDAR only captures points on object surfaces, 3D object centers are likely to reside on the empty locations with zero features, which is unfriendly to the current designs of detection heads.~\cite{pointpillar, centerpoint}.
So we add two $3 \times 3$ convolutions to fill most of the holes on the object centers.

As for the detection head and loss function, we adopt the same settings as PointPillars~\cite{pointpillar} for simplicity.
Specifically, we use the SSD~\cite{ssd} head, the smooth L1 bounding box localization loss $\mathcal{L}_{loc}$, the classification loss $\mathcal{L}_{cls}$ in the form of focal loss~\cite{focalloss}, and the direction loss $\mathcal{L}_{dir}$ penalizing wrong orientations. The final loss function is Eq~\ref{eq:loss}, where $N_p$ is the number of positive samples. We leave the detailed setting in supplementary materials.

\begin{equation}
\begin{aligned}
\label{eq:loss}
    \mathcal{L}=\frac{1}{N_{p}}(\beta_{loc}\mathcal{L}_{loc}+\beta_{cls}\mathcal{L}_{cls}+\beta_{dir}\mathcal{L}_{dir})
\end{aligned}
\end{equation}
\subsection{Two Stage SST}
Although our main contribution lies in the design of the single stride architecture in the first stage, there is a considerable gap between the single stage detector and the two stage detector. To match the performance with current two stage detectors, we apply LiDAR-RCNN~\cite{lidarrcnn} as our second stage. LiDAR-RCNN is a lightweight second stage network consists of a simple PointNet~\cite{pointnet} for feature extraction, only taking the raw point cloud inside proposal as input.
\subsection{Discussion}
Because of the distinctions between point clouds and RGB images, there are several differences in the design choices and motivations between our design and Swin-Transformer~\cite{swin} as highlighted here. 
\begin{itemize}
    \item Our SST network follows the \textbf{single-stride} guideline, while Swin-Transformer follows the hierarchical structure with \textbf{multi-stride}, which uses ``token merge'' to increase the receptive field.
    \item The tokens for our region-based attention scatter \textbf{sparsely} because of the sparsity of point clouds, while the tokens in vision transformers have \textbf{dense} layouts. This is one of the reasons for the efficiency of SST even in the single stride architecture.

\end{itemize}

\section{Experiments}
\subsection{Dataset}
We conduct our experiments on Waymo Open Dataset (WOD)~\cite{wod}.
The dataset contains 1150 sequences in total (more than 200K frames), 798 for training, 202 for validation and 150 for test.
Each frame covers a scene with a size of $150m \times 150m$.
It is a very challenging dataset and adopted as the benchmark in many recent state-of-the-art methods.
\subsection{Implementation Details}
We implement our model based on the popular 3D object detection codebase -- MMDetection3D\cite{mmdet3d}, which provides standard and solid baselines.
Please refer to supplementary materials for more details.
\mypar{Model Setup}
For generality, we build our \emph{Single-stride Sparse Transformer} (SST) on the basis of popular PointPillars~\cite{pointpillar}. We replace its backbone with 6 consecutive \emph{Sparse Regional Attentions} (SRA) blocks, and each block contains 2 attention modules as Fig. \ref{fig:arch} shows. All the attention modules are equipped with 8 heads, 128 input channels, and 256 hidden channels. In \emph{Regional Grouping}, each region covers a volume with size $3.84m \times 3.84m \times 6m$. As for other parts, SST follows the implementation of PointPillars in MMDetection3D. We use the BEV pillar size of $0.32m \times 0.32m \times 6m$, which can be easily extended to the 3D voxels with smaller heights. 
\mypar{Model Variants} We develop several variants of SST in our experiments. SST\_1f: basic single-stage model using 1-frame point cloud. SST\_3f: consecutive 3 frame point clouds are used as model input, and the point cloud in different frames are concatenated together after aligning the ego-pose. SST\_TS\_1f and SST\_TS\_3f: two stage model based on above models, using a standard LiDAR-RCNN~\cite{lidarrcnn} for refinement.
\mypar{Training Scheme} We train our model for 24 epochs (2$\times$) on WOD with AdamW optimizer and cosine learning rate scheduler. The maximum learning rate is $0.001$, and the weight decay is $0.05$. 
\subsection{Comparison with State-of-the-art Detectors}
We compare our SST with state-of-the-art methods in Table~\ref{tab:sota_veh_small} (vehicle) and Table~\ref{tab:sota_ped_small} (pedestrian). We divide current methods into the branches of one-stage and two-stage detectors for fair comparison.
\par
Table \ref{tab:sota_veh_small} shows the results on vehicles, where our models achieve competitive performances.
With a lightweight second stage for refinement, our two-stage detectors are comparable with state-of-the-art methods.
\par
Table \ref{tab:sota_ped_small} shows the results on pedestrians. Due to the tiny size and non-rigid property, pedestrian detection is more challenging than vehicle detection. Networks are prone to confuse pedestrians with other slim objects, like poles and trees, leading to a high false positive rate. Under such cases, our best model \textbf{outperforms all other methods in the challenging pedestrian class}. SST\_TS\_3f is \bb{4.4} AP ahead of the second best RSN with the same temporal information (3 frames). We owe such leading performance to the single-stride characteristic of SST. 
\begin{table}[ht]
\small
\centering
\setlength{\tabcolsep}{3.5mm}{

\resizebox{0.95\linewidth}{!}{
\begin{tabular}{l|c|c}
  \specialrule{1pt}{0pt}{1pt}
\toprule
\multirow{2}{*}{Methods} & \multirow{1}{*}{LEVEL\_1} & \multirow{1}{*}{LEVEL\_2}\\
 &   3D AP/APH           &       3D AP/APH       \\
\midrule
\textit{One-Stage Methods} \\
\midrule
SECOND \ddag~\cite{second} & 72.27/71.69 & 63.85/63.33\\
MVF~\cite{MVF}  & 62.93/-      & -/-   \\
LaserNet \P~\cite{lasernet}  & 56.10/-  & -/48.40 \\
AFDet~\cite{afdet}  & 63.69/-  & -/- \\
Pillar-OD~\cite{pillarbased}     & 69.80/-   & -/-  \\
PPC~\cite{tothepoint}  & 65.2/- & -/56.7\\
VoTr-SSD~\cite{votr}    &  68.99/68.39  &  60.22/59.69   \\
RangeDet~\cite{rangedet} & 72.85/72.33 & 64.03/63.57\\
CenterPoint-Voxel~\cite{centerpoint} & \second{74.78}/\second{74.22} & \second{66.70}/\second{66.19}\\
PointPillars$^\ast$~\cite{pointpillar}   & 72.08/71.53  & 63.55/63.06 \\
{SST\_1f (Ours)} & {74.22/73.77} & {65.47/65.07}\\
{SST\_3f (Ours)} & \best{77.04}/\best{76.56} & \best{68.50}/\best{68.08}\\
\midrule
\textit{Two-Stage Methods} \\
\midrule
Voxel RCNN~\cite{voxelrcnn}  & 75.59/-  & 66.59/-\\
RCD~\cite{rcd} & 69.0/68.5  & -/- \\
VoTr-TSD~\cite{votr}   & 74.95/74.25 & 65.91/65.29 \\
LiDAR-RCNN~\cite{lidarrcnn} & 76.0/75.5 & 68.3/67.9\\
Pyramid RCNN~\cite{pyramidrcnn} & 76.30/75.68 & 67.23/66.68\\
Voxel-to-Point~\cite{voxeltopoint} & 77.24/- & 69.77/- \\
3D-MAN~\cite{3dman} & 74.53/74.03 & 67.61/67.14 \\
Part-A2-Net \ddag ~\cite{parta2} & 77.05/76.51 & 68.47/67.97\\
CenterPoint-Pillar~\cite{centerpoint} & 76.10/75.50 & 68.00/67.50\\
CenterPoint-Voxel~\cite{centerpoint} & 76.59/7605 & 68.85/68.35\\
PV-RCNN~\cite{pvrcnn}   & 77.51/76.89     & 68.98/68.41 \\
PV-RCNN++~\cite{pvrcnnpp}   & \best{78.79}/\best{78.21}     & \best{70.26}/\best{69.71}\\
RSN\_1f \dag~\cite{rsn} & 75.10/74.60 &66.00/65.50\\
RSN\_3f \dag~\cite{rsn} & {78.40/78.10} & {69.50/69.10}\\
{SST\_TS\_1f (Ours)} & 76.22/75.79 & 68.04/67.64\\
{SST\_TS\_3f (Ours)} & \second{78.66}/\second{78.21} & \second{69.98}/\second{69.57}\\
\bottomrule
  \specialrule{1pt}{1pt}{0pt}
\end{tabular}
}
}
\caption{
    Performances of \textbf{vehicle} detection on the Waymo Open Dataset validation split.
    We mark the best result in \best{red}, and the second result in \second{blue}.
    \dag:  RSN~\cite{rsn} is not a typical two stage detector, we put it here because it uses a segmentation network to remove background first.
    $^\ast$: re-implemented by MMDetection3D.
    \P: from~\cite{tothepoint}.
    \ddag: from \cite{pvrcnnpp}.} \label{tab:sota_veh_small}
\end{table}

\begin{table}[ht]
\small
  \centering
\resizebox{0.95\linewidth}{!}{
\setlength{\tabcolsep}{3.5mm}{
\begin{tabular}{l|c|c}
  \specialrule{1pt}{0pt}{1pt}
\toprule
\multirow{2}{*}{Methods} & \multirow{1}{*}{LEVEL\_1} & \multirow{1}{*}{LEVEL\_2}\\
 &   3D AP/APH      &   3D AP/APH\\
\midrule
\textit{One-Stage Methods} \\
\midrule
LaserNet\P~\cite{lasernet}   & 62.9/-  & -/45.4\\
SECOND \ddag~\cite{second} & 68.70/58.18 & 60.72/51.31 \\
MVF~\cite{MVF}  & 65.33/-      & -/-  \\
Pillar-OD~\cite{pillarbased}     & 72.51/-   & -/- \\
PPC~\cite{tothepoint}  & 73.90/- & -/59.60 \\
RangeDet~\cite{rangedet} & {75.94}/\second{71.94} & 67.60/\second{63.89} \\
CenterPoint-Voxel~\cite{centerpoint} & 75.82/{69.65} & {68.34/62.62} \\
PointPillars$^{\ast}$~\cite{pointpillar}   & 70.59/56.70  & 62.84/50.25\\
{SST\_1f (Ours)} & \second{78.71}/69.55 & \second{70.02}/61.67 \\
{SST\_3f (Ours)} & \best{82.42}/\best{77.96} & \best{75.14}/\best{70.88} \\
\midrule
\textit{Two-Stage Methods} \\
\midrule

LiDAR-RCNN~\cite{lidarrcnn} & 71.2/58.7 & 63.1/51.7\\
3D-MAN~\cite{3dman} & 71.71/67.74 & 62.58/59.04\\
Part-A2-Net \ddag~\cite{parta2} & 75.24/66.87 & 66.18/58.62\\
PV-RCNN~\cite{pvrcnn} & 75.01/65.65  &  66.04/57.61\\
PV-RCNN++~\cite{pvrcnnpp} & 76.67/67.15  &  68.51/59.72 \\
CenterPoint-Pillar~\cite{centerpoint} & 76.10/65.10 & 68.10/57.90\\
CenterPoint-Voxel~\cite{centerpoint} & 79.02/73.44 & {70.98}/65.75\\
RSN\_1f \dag~\cite{rsn} & 77.80/72.70 & 68.30/63.70 \\
RSN\_3f \dag~\cite{rsn} & {79.40}/\second{76.20} & 69.90/\second{67.00} \\
{SST\_TS\_1f (Ours)} & \second{81.39}/{74.05} & \second{72.82}/{65.93} \\
{SST\_TS\_3f (Ours)} & \best{83.81}/\best{80.14} & \best{75.94}/\best{72.37} \\
\bottomrule
  \specialrule{1pt}{1pt}{0pt}
\end{tabular}
}
}
\caption{Performance of \textbf{pedestrian} detection on the Waymo Open Dataset official validation split. Please refer to Table \ref{tab:sota_veh_small} for the meanings of the notions in this table.} \label{tab:sota_ped_small}
\end{table}
\begin{figure*}[!t]
    \centering
    \includegraphics[width=2.0\columnwidth]{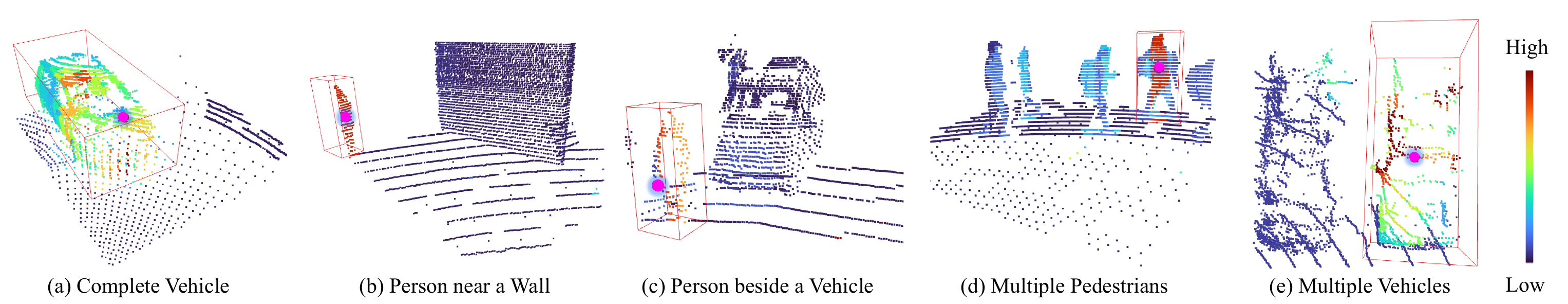}
    \caption{Visualization of the learned sparse regional attention. Each figure shows the attention weight distribution between the query token (pink dot) and all other tokens in the local region. Best viewed in color.}
    \label{fig:attention}
\end{figure*}
\subsection{Deep Investigation of Single Stride}

\mypar{Single-stride models better use dense observations.}  
\textbf{First}, SST has more advantages in short-range metrics (0m - 30m) than in long-range metrics (50m - inf): In Table~\ref{tab:ped_with_distance}, SST\_1f outperforms the PointPillars counterpart in short-range metric by \textbf{12.8} AP for pedestrian class, but the margin is not that significant over PointPillars in the long-range metric. 
\textbf{Second}, SST benefits more from multi-frame data. In Table~\ref{tab:ped_with_distance}, RSN~\cite{rsn} got improved by \bb{6.4} AP in long-range metric from RSN\_1f to RSN\_3f, while the performance of SST in long-range metric gets more significantly improved by \bb{10.4} AP from SST\_1f to SST\_3f. 
\begin{table}[h]
\small
\begin{center}
\resizebox{0.9\linewidth}{!}{
\begin{tabular}{l@{\ \ }c@{\ \ }c@{\ \ }c@{\ \ }c@{\ \ }c@{\ \ }}
  \toprule 
   \multirow{2}{4em}{Method} &  \multicolumn{4}{c}{LEVEL\_1 Pedestrian AP}\\
   & Overall & 0-30m & 30-50m & 50m-inf\\

       \cmidrule(r){1-2}
    \cmidrule(r){3-5}

   RSN\_1f            & 77.8 & 83.9 & 74.1 & 62.1 \\
   RSN\_3f            & 79.0 & 84.5 & 78.1 & 68.5 \\ \hline
   PointPillars       & 70.6 & 72.5 & 71.9 & 63.8\\   
   PointPillars\_3f   & 73.7 & 72.9 & 75.9 & 70.6\\   \hline
   SST\_1f            & 78.7\up{8.1} & 85.3\up{12.8} & 77.0\up{5.1} & 63.4\down{0.4}\\ 
   SST\_3f            & 82.4\up{8.7} & 86.1\up{13.2} & 81.2\up{5.3} & 73.8\up{3.2}\\ 
  \bottomrule
 
 \end{tabular}
 }
\end{center}
\vspace{-3mm}
\caption{Distance-conditioned \bb{pedestrian} detection performance. Our SST has larger advantages in short-range metrics and the multi-frame setting, where points are more dense. The increases and decreases are calculated based on PointPillars baseline.}
\label{tab:ped_with_distance}
\end{table}
\mypar{Does the single stride model fail on large vehicles? } 
As smaller strides reduce the receptive fields, it would be a major concern whether our model has sufficient receptive fields for extreme cases, \eg, extremely large vehicles. 
We therefore divide all the vehicles into three groups according to the lengths of their ground-truth boxes, and evaluate the recalls of SST on them. Please refer to supplementary materials for the evaluation details.
In Table~\ref{tab:diff_length}, our SST outperforms the PointPillars baseline for all vehicles, even those longer than $8m$.
This supports that our attention mechanism provides proper receptive fields in the single stride architecture.
\begin{table}[h]
\small
\begin{center}
\resizebox{0.8\linewidth}{!}{
\begin{tabular}{l@{\ \quad}l@{ \quad}l@{ \quad}l@{\ }}
  \specialrule{0.7pt}{0pt}{1pt}
  \specialrule{0.7pt}{0pt}{2pt}
  \multirow{2}{4em}{Methods} &  \multicolumn{3}{c}{Vehicle Recalls (IoU=0.7)}\\
  & $[0m, 4m]$ & $[4m, 8m]$ & $[8m, +\infty]$\\
    \specialrule{0.5pt}{1pt}{2pt}
PointPillars & 40.60 & 73.11 & 10.59 \\
    \specialrule{0pt}{1pt}{1pt}
SST\_1f & 41.31 \up{0.71} & 80.85 \up{7.74} & 13.41 \up{2.82} \\

  \specialrule{0.7pt}{2pt}{1pt}
  \specialrule{0.7pt}{0pt}{1pt}
 
 \end{tabular}
}
\end{center}
\vspace{-3mm}
\caption{Recalls for vehicles with different lengths. The vehicles with lengths in $[0m, 4m]$ and $[8m, +\infty]$ are rare (7.3\% and 1.6\% in WOD) and hardly get sufficient training, so their performances are relatively low. 
}
\label{tab:diff_length}
\end{table}
\mypar{Localization quality test with stricter IoU thresholds.}
By preserving the original resolution, our SST is supposed to localize objects more precisely as in~\cite{cornernet}. 
To verify this, we evaluate SST with higher 3D IoU thresholds (0.8 for vehicle, 0.6 for pedestrian). 
In Table~\ref{tab:high_iou}, we compare our models with the PointPillars baseline and other models with available results from~\cite{offboard}, then a couple of interesting findings emerge: 
\begin{enumerate}
    \item Comparing MVF++~\cite{offboard} with our SST\_1f on vehicles, MVF++ is slightly better than SST\_1f under the normal threshold, while SST\_1f is better with the stricter threshold. This suggests the single stride structure enables more precise localization of vehicles.
    \item The 3DAL~\cite{offboard} is an offboard method using all the past and future frames in a sequence (around 200 frames) and is equipped with tracking~\cite{3dmotbaseline}. Nonetheless, our best model \textit{SST\_TS\_3f surprisingly surpasses 3DAL on pedestrian} on both IoU thresholds with as few as 3 frames of point clouds.
\end{enumerate}
These findings suggest that the single-stride architecture is capable of better localizing objects with full and fine-grained information.
\begin{table}[h]
\begin{center}
\resizebox{0.95\linewidth}{!}{
\begin{tabular}{@{}l@{\ \ }c@{\ \ }c@{\ \ }c@{\ \ }c@{\ \ }c@{\ \ }}
  \toprule 
   \multirow{2}{4em}{Method} & \multirow{2}{4em}{Frames} &  \multicolumn{2}{c}{Vehicle} & \multicolumn{2}{c}{Pedestrian}\\
   & & Normal & Strict & Normal & Strict\\

       \cmidrule(r){1-2}
    \cmidrule(r){3-6}

  \textit{ Single frame } \\ 
  PointPillars~\cite{pointpillar}     &    1     & 72.08 & 36.83 & 70.59 & 44.86\\  
  PV-RCNN$^\ast$~\cite{pvrcnn}         &    1     & 70.47 & 39.16 & 65.34 & 45.12\\ 
  MVF++$^\ast$~\cite{offboard}         &    1     & \textbf{74.64} & 43.30 & 78.01 & {56.02} \\ 
  SST\_1f (Ours)                      &    1     & 74.22 & \textbf{44.08} & \textbf{78.71} & \bb{56.12}\\ 
  \midrule

  \textit{ Multiple frames } \\
  MVF++ w. TTA$^\ast$~\cite{offboard}  &    5     & 79.73 & 49.43 & 81.83 & 60.56\\ 
  3DAL$^\ast$~\cite{offboard}          &   all\dag& \textbf{84.50} & \textbf{57.82} & 82.88 & 63.69\\ 
  SST\_TS\_3f (Ours)                  &    3     & 78.66 & 49.35 & \textbf{83.81} & \textbf{65.06}\\ 
  \bottomrule
 
 \end{tabular}
}
\end{center}
\vspace{-3mm}
\caption{Localization quality test with stricter IoU threshold. The normal and strict thresholds for vehicles are 0.7 and 0.8, and are 0.5 and 0.6 for pedestrians. $\ast$: the results are from~\cite{offboard}. TTA: test-time data augmentation. \dag: the offline setting using all the past and future frames in a point cloud sequence.}
\label{tab:high_iou}
\end{table}
\mypar{Comparison with other alternatives.}
There are some potential alternatives to our SST in order to preserve the input resolution. Here we make a comprehensive comparison. We first introduce these alternative models as follows.
\textbf{PointPillars-SS}: The single stride version of PointPillars introduced in Sec.~\ref{sec:pilot}.
\textbf{SparsePillars-SS}: We replace all the standard 2D convolutions in backbone of PointPillars-SS with Submanifold Sparse Convolutions~\cite{sparseconv, second}. Due to the sparsity, SparsePillars-SS also faces the issue of ``empty hole'' (details in Sec.~\ref{sec:method_integrate_detection}) as in SST, so we add two more 2D convolutions before its detection head.
\textbf{HRNetV2p-W18}~\cite{hrnet}: HRNet maintains the high resolution while building multi-scale features.
We adopt the standard HRNetV2p-W18 from MMDetection~\cite{mmdet} for the experiment. To keep the output resolution in HRNet the same as PointPillars, we reduce the stride of the first two convolutions in HRNet from 2 to 1. 
All the alternatives have the same setting with SST\_1f except their backbones.
Table~\ref{tab:comp_with_ss_pp} shows the comparison between different models.
\begin{table}[h]
\begin{center}
\resizebox{0.99\linewidth}{!}{
\begin{tabular}{l@{\ }c@{ }c@{\quad }c@{\ }c@{\ }c@{\ }}
  \specialrule{0.7pt}{0pt}{1pt}
  \specialrule{0.7pt}{0pt}{2pt}
  \multirow{2}{4em}{Models} & \multirow{2}{4em}{Vehicle 3D AP} & \multirow{2}{4em}{Pedestrian 3D AP} & \multirow{2}{4em}{\#param.}& \multirow{2}{4em}{Latency (ms)}& \multirow{2}{4em}{Memory (GB)}\\
  &&& \\
    \specialrule{0.5pt}{1pt}{2pt}
PointPillars     & 64.01  & 60.85 & 6.4M & \bf{60} & \bf{5.4}\\
    \specialrule{0pt}{1pt}{1pt}
PointPillars-SS  & 64.69  & 64.32 & 6.4M  & 185 & 8.5\\
    \specialrule{0pt}{1pt}{1pt}
SparsePillars-SS  & 51.57  & 61.55 & 6.4M  & 67 & 5.8 \\
    \specialrule{0pt}{1pt}{1pt}
SparsePillars-SS$_{5\times 5}$\dag  & 55.40  & 61.28 & 17.1M  & 81 & 5.9 \\
SparsePillars-SS$_{7\times 7}$\dag  & 56.77  & 60.87 & 33.9M  & 97 & 5.9 \\
    \specialrule{0pt}{1pt}{1pt}
HRNetV2p-W18~\cite{hrnet} & 64.38 & 61.09 &  26.2M & 130 & 7.6 \\
    \specialrule{0pt}{1pt}{1pt}
SST\_1f          & \bf{67.86}  & \bf{70.94} & \bf{1.6M} & 97 & 6.8\\
    \specialrule{0pt}{1pt}{1pt}

  \specialrule{0.7pt}{2pt}{1pt}
  \specialrule{0.7pt}{0pt}{1pt}
 
 \end{tabular}
 }
\end{center}
\vspace{-3mm}
\caption{Comparison with alternatives to SST. Using 20\% data for training. The latency is evaluated with standard benchmarking script in MMDetection3D on 2080Ti GPUs. \dag: The size of all kernels in sparse convolution increases to $5 \times 5$ or $7 \times 7$.}
\label{tab:comp_with_ss_pp}
\end{table}
\par
In Table~\ref{tab:comp_with_ss_pp}, our method outperforms all other alternatives with relatively low latency. Besides, two things need to be noticed:
(1) \textbf{SparsePillars-SS} is much worse than other models in vehicle class. Due to the properties of submanifold sparse convolution, this model suffers from more severe receptive field shrinkage than PointPillars-SS. For example, if a vehicle part is isolated with all the surrounding voxels being in empty, it can not perceive information from other parts in the whole forward process. On the contrary, the attention mechanism in SST well addresses this issue while maintaining sparsity. (2) \textbf{HRNetV2p-W18} allocates too much computation on the high-stride (low resolution) branches which is not needed in 3D object detection. So the capacity of its high-resolution branch is limited, leading to its inferior performance.
\subsection{Qualitative Analysis of Sparse Attention}
We visualize the attention weights in Fig.~\ref{fig:attention} and list our observations as follows.
\mypar{Sufficient Coverage}
In Fig.~\ref{fig:attention} (a) \emph{Complete Vehicle}, the query token (pink dot) in the car has strong relation with all other parts of the car. In other words, this single token can effectively cover the whole car. This demonstrates that the attention mechanism is indeed effective to enlarge the receptive field.
\mypar{Semantic Discrimination} In Fig.~\ref{fig:attention} (b) \emph{Person near a Wall}, the query token on the person builds strong dependency with other body parts, but has little relations with background points, \eg, wall. In Fig.~\ref{fig:attention} (c) \emph{Person beside a Vehicle}, the pedestrian standing next to the vehicle attends only with itself. These two cases reveal that the learned sparse attention weight is discriminative between different semantic classes. This property helps distinguish pedestrians from other slim objects and reduces false positives.
\mypar{Instance Discrimination}
In the crowded cases,  such as Fig.~\ref{fig:attention} (d) \emph{Multiple Pedestrians}, the query token in a person mainly focuses on the same person. Due to the high semantic similarity, it also slightly attends to other people. In Fig.~\ref{fig:attention} (e) \emph{Multiple Vehicles}, the query token in the vehicle almost has no dependency on the nearby vehicles. These two cases suggest that the learned sparse attention weights are discriminative for different instances.
\subsection{Hyper-parameter Ablation}
\mypar{Region Size} We show the performance under different region sizes for \emph{Regional Grouping} in Table~\ref{tab:region_size}. SST is in general robust to the region size and slightly better with larger regions. Especially, SST has the best performance in pedestrian detection with the largest local region size.
It suggests that the local context is helpful to recognize pedestrians.
For example, pedestrians are more likely to appear on the sidewalks than on vehicle lanes.
\begin{table}[h]
\small
\begin{center}
\begin{tabular}{l@{\ \ }c@{\ }c@{\ \quad}c@{\ \ }}
  \toprule 
   \multirow{2}{6em}{Region Size} &\multirow{2}{6em}{Max number of voxels} &  \multicolumn{2}{c}{LEVEL\_1 AP/APH}\\
    & & Vehicle & Pedestrian \\
   \midrule

  $3.20m$  &   \hspace{-3mm}100   & 66.9/66.4 & 70.4/56.9\\ 
  $3.84m$  &   \hspace{-3mm}144   & \textbf{67.9/67.3} & 70.9/\textbf{57.3}\\ 
  $4.48m$  &   \hspace{-3mm}196   & 67.8/67.3 & 70.6/56.5\\ 
  $5.12m$  &   \hspace{-3mm}256   & 66.9/66.3 & \textbf{71.1}/57.1\\ 
  \bottomrule
 
 \end{tabular}
\end{center}
\caption{Ablation of the region size. Using 20\% data for training.}
\label{tab:region_size}
\end{table}
\mypar{Network Depth}
SST is relatively shallow by design thanks to the large receptive fields from the attention mechanism. In Table~\ref{tab:depth} we show the impact of model depths on SST. In general, SST is robust to different depths, and the performance of pedestrian class is even slightly better with fewer layers. This demonstrates our method does not rely on a very large or deep model, thus it is easier to build efficient single-stride models.
\begin{table}[h]
\small
\begin{center}
\begin{tabular}{l@{\ \ }c@{\ \quad}c@{\ \ }}
  \toprule 
   \multirow{2}{6em}{Number of SRA blocks} &  \multicolumn{2}{c}{LEVEL\_1 AP/APH}\\
    & Vehicle & Pedestrian \\
   \midrule


  5  &    67.7/67.1 & \textbf{71.1/57.9}\\ 
  6  &    \textbf{67.9/67.3} & 70.9/57.3\\ 
  7  &    67.8/67.3 & 70.6/56.1\\ 
  \bottomrule
 
 \end{tabular}
\end{center}
\vspace{-3mm}
\caption{Ablation of the network depth. Using 20\% training data.}
\label{tab:depth}
\end{table}
\section{Conclusion and Limitations}
In this paper, we analyze the impact of the network stride on 3D object detectors for autonomous driving, and empirically show that 3D object detectors do not really need downsampling. To build a single-stride network, we adopt the sparse regional attention to address the problem of insufficient receptive fields and avoid expensive computation. By stacking the sparse attention modules, we propose the Single-stride Sparse Transformer, achieving state-of-the-art performance on the Waymo Open Dataset. Due to the single stride structure, our models obtain remarkable performance on the challenging pedestrian class. Without elaborated optimization, our model uses slightly more memory than baseline models, and we will pursue a more memory-friendly model in the future. We wish our work could break the stereotype in the backbone design of point cloud data, and inspire more thoughts on the specialized architectures.

{\small
\bibliographystyle{ieee_fullname}
\bibliography{egbib}
}
\clearpage
\section*{\LARGE Appendices}
\setcounter{table}{0}
\setcounter{figure}{0}
\setcounter{equation}{0}
\renewcommand{\thetable}{\Alph{table}}
\renewcommand{\thefigure}{\Alph{figure}}
\renewcommand{\theequation}{\Alph{equation}}
\appendix
\section{Submission on Test Server}
Due to the submission frequency limit of the Waymo test server, we only report the results of our best model. We compare SST with the three most competitive methods and report their performances in the multi-frame setting from the official leaderboard.
The results are shown in Table~\ref{tab:test_veh} and Table~\ref{tab:test_ped}.
The performance of SST on vehicle class is comparable with these methods, and the performance of SST on pedestrian class significantly outperforms other methods.

\begin{table}[h]
\small
\centering
\setlength{\tabcolsep}{3.5mm}{

\resizebox{0.95\linewidth}{!}{
\begin{tabular}{l|c|c}
  \specialrule{1pt}{0pt}{1pt}
\toprule
\multirow{2}{*}{Methods} & \multirow{1}{*}{LEVEL\_1} & \multirow{1}{*}{LEVEL\_2}\\
 &   3D AP/APH           &       3D AP/APH       \\
\midrule
PVRCNN\_2f~\cite{pvrcnn} & 81.06/80.57 & 73.69/73.23 \\
CenterPoint\_2f~\cite{centerpoint} & 81.05/80.59  & 73.42/72.99 \\
RSN\_3f~\cite{rsn} & 80.70/80.30 & 71.90/71.60\\
{SST\_TS\_3f (Ours)} &  80.99/80.62 & 73.08/72.74 \\
\bottomrule
  \specialrule{1pt}{1pt}{0pt}
\end{tabular}
}
}
\caption{Performance of \textbf{vehicle} detection on test split of Waymo Open Dataset.}
\label{tab:test_veh}
\end{table}

\begin{table}[h]
\small
\centering
\setlength{\tabcolsep}{3.5mm}{

\resizebox{0.95\linewidth}{!}{
\begin{tabular}{l|c|c}
  \specialrule{1pt}{0pt}{1pt}
\toprule
\multirow{2}{*}{Methods} & \multirow{1}{*}{LEVEL\_1} & \multirow{1}{*}{LEVEL\_2}\\
 &   3D AP/APH           &       3D AP/APH       \\
\midrule
PVRCNN\_2f~\cite{pvrcnn} & 80.31/76.28 & 73.98/70.16\\
CenterPoint\_2f~\cite{centerpoint} & 80.47/77.28  & 74.56/71.52 \\
RSN\_3f~\cite{rsn} & 78.90/75.60 & 70.70/67.80\\
{SST\_TS\_3f (Ours)} &  83.05/79.38 & 76.65/73.14 \\
\bottomrule
  \specialrule{1pt}{1pt}{0pt}
\end{tabular}
}
}
\caption{Performance of \textbf{pedestrian} detection on test split of Waymo Open Dataset.}
\label{tab:test_ped}
\end{table}
\section{Discussion of Sparse Operations}
Due to the space limit of the main paper, we leave the discussion on sparse operations in the supplementary materials.
In this section, we discuss two problems for sparse operations: (1) \textit{insufficient receptive field of submanifold sparse convolution (SSC)}~\cite{sparseconv}, and (2) \textit{the difficulties of downsampling/upsampling in sparse data}.
\subsection{Insufficient Receptive Field of Submanifold Sparse Convolution (SSC)}
\begin{figure*}[!t]
	\centering
	\includegraphics[width=1.99\columnwidth]{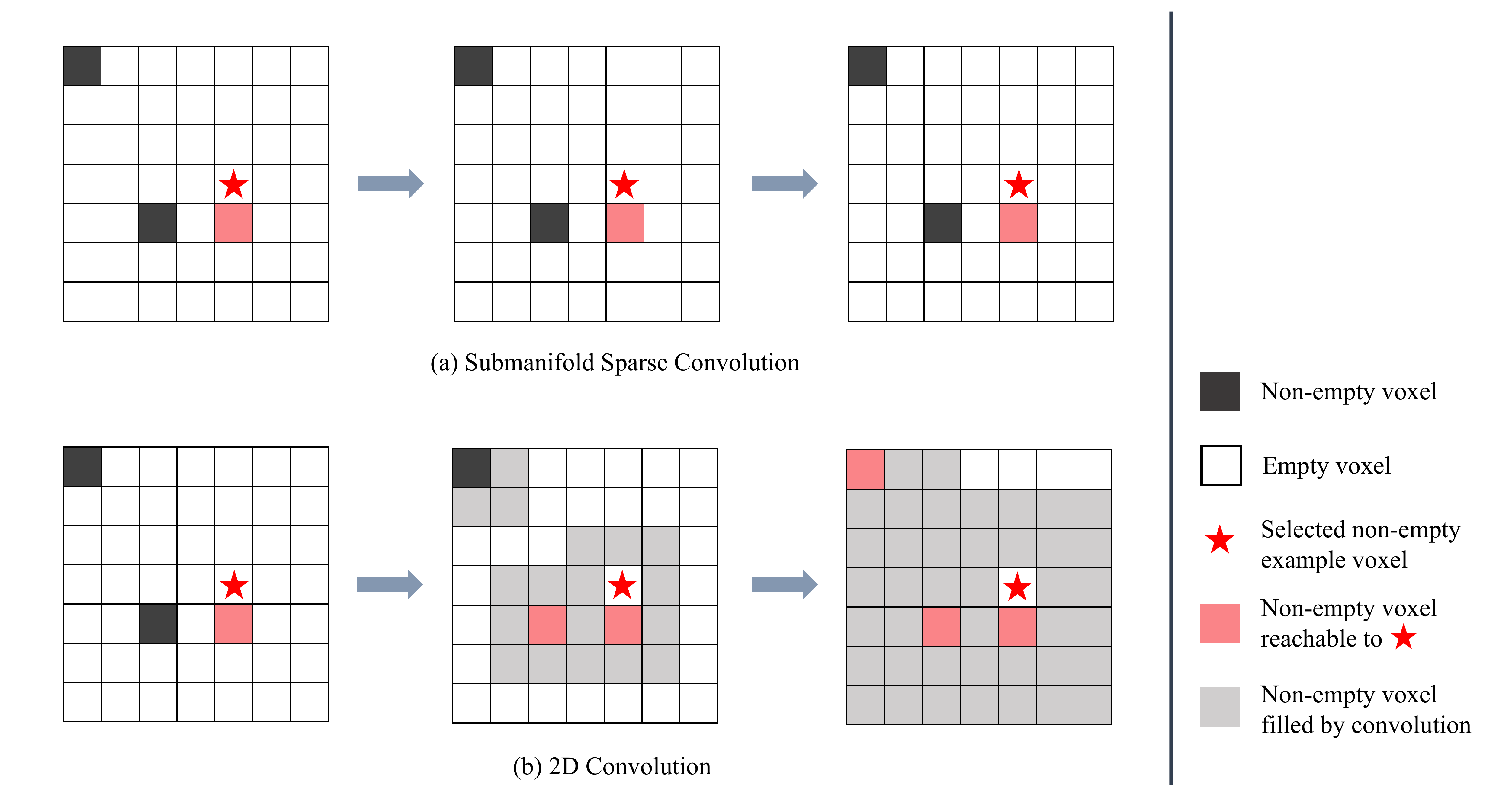}
	\caption{Illustration of receptive enlarging in the $3 \times 3$ submanifold sparse convolution (SSC) and the standard $3 \times 3$ convolution. In SSC, only the information of the voxel (the pink one) covered by the kernel can reach to the red star. In 2D convolution, all non-empty voxels can reach to the red star after 2 convolution layers, because the empty locations are ``filled'' by the convolution. }
	\label{fig:supp_spconv1}
\end{figure*}

In Sec. \blue{1} and Table \blue{7} in our main paper, we briefly point out that the SSC-based single-stride architecture faces a severe problem of the insufficient receptive field.
We demonstrate this issue here in Fig.~\ref{fig:supp_spconv1} by comparing the behaviors of SSC and standard 2D convolution in sparse data. Both the SSC and standard convolutions have two layers with a kernel size of three. However, the SSC could not reach the voxel on the top-left corner from the voxel marked with a star, while the standard convolution is capable of doing this. This example intuitively illustrates the insufficiency of receptive fields for SSC, and we explain the reasons in detail as follows.

The SSC do not ``fill'' empty voxels for the sake of efficiency, which largely constrains the information communication between voxels.
Under such conditions, in Fig.~\ref{fig:supp_spconv1} (a), only one voxel (the pink one) in has information communication with the one marked by a red star if the kernel size is $3\times 3$.
On the contrary, Fig.~\ref{fig:supp_spconv1} (b) shows that the 2D convolution can gradually enlarge the receptive field by involving the empty voxels in the convolution process, which is more effective for aggregating information compared to the SSC.

To give an experimental illustration, we conduct experiments on the class of vehicles, which require sufficient receptive field for detection. In the Table~\blue{7} of the main paper, replacing the $3\times 3$ standard convolutions with SSC will cause a significant drop of AP from 64.69 to 51.57. We further increase the receptive field by expanding the kernel size of SSC to $5\times 5$ and $7\times 7$. These improve the performance from the 3D AP 51.57 to 55.40 and 56.77, but there is still a large gap to the variant using standard convolutions. Therefore, these numbers support our analyses on the insufficient receptive fields of SSC.

\subsection{Downsampling/Upsampling in Sparse Data}
Although downsampling and upsampling are common in dense data, \eg, pooling in CNN, token merge in Swin-Transformer, it is non-trivial to transfer these techniques to sparse data like point clouds.
A variant of SSC named Sparse Convolution (SC) follows the standard convolution to implement the downsampling and upsampling in sparse data. With such implementation, data loses sparsity rapidly~\cite{second, centerpoint} and this leads to high computational overhead.
\par
In our sparse Transformer, downsampling/upsampling by token merge~\cite{swin} also needs careful consideration. First, the downsampling operation is still an open problem for point clouds: what is the best way to merge the varied number of tokens scattered in different spatial locations?
Second, the upsampling operation is also non-trivial and requires future research: how to recover a couple of tokens in different locations from a single token effectively and efficiently?
In developing the SST, we encounter these challenges and find it difficult to offer satisfying solutions. Although we have bypassed these difficulties by adopting the single-stride architecture, we hope future research may work on this downsampling/upsampling question and better utilizes sparse data.

\section{Potential Improvements}
In order to rule out unimportant factors and present a clean architecture, we only inherit the basic framework of PointPillars~\cite{pointpillar}. So there is a large room for further performance improvements, and we list some of them as follows. We will adopt these techniques in our future work.
\mypar{IoU Prediction.} In detection, the classification score of a bounding box are not always consistent with the real regression quality.
So many recent methods~\cite{parta2, pvrcnn, centerpoint, rangedet} use another branch to predict the IoU between output bounding boxes and the corresponding ground-truth boxes, and use the predicted IoU to correct the classification scores.

\mypar{More Powerful Second Stage.}
We use LiDAR-RCNN~\cite{lidarrcnn} as our second stage, which is a lightweight PointNet-like module only takes the raw point cloud as input.
So it has no effect on our first stage and is convenient for our analysis of single-stride architecture.
However, its performance is inferior to some other elaborately designed RCNNs, \eg, CenterPoint~\cite{centerpoint}, PartA2~\cite{parta2}, PVRCNN~\cite{pvrcnn}, PyramidRCNN~\cite{pyramidrcnn}, which reuse the features from the single stage to achieve better refinement.
With the point-level features interpolated from feature maps in the first stage, SST can be equipped with most of these methods and aim for better abilities.

\mypar{Incorporating Advanced Techniques in Vision Transformer.}
We have witnessed the fast progress of vision transformers. Many advanced techniques can be borrowed to enhance the performance of SST. \textbf{(1) Better efficiency}: There are a lot of techniques can be adopted to improve our efficiency, for example, token selection~\cite{pnpdetr, dynamicvit}, attention simplication~\cite{linformer}. \textbf{(2) Better efficacy}: Some techniques can be used to make SST more effective, \eg, relative positional encoding~\cite{rpe}, different attention mechanism~\cite{xcit}.
\section{Computational Complexity Compared with Convolutions}

We investigate the computational complexity of the SST architecture and convolutional architectures. Our analyses demonstrate that SST has a unique advantage in efficiency by utilizing the sparsity of point clouds and the regional grouping.

Following the calculation in Swin-Transformer~\cite{swin}, we inspect the computational complexities of convolutional architectures and SST. For an input scene size of $h\times w$, a convolution layer with kernel size $k \times k$ and channel number $C$ has the complexity as Equation~\ref{eq:conv}. On the same scene, an SRA operation has the complexity as Equation~\ref{eq:sra}, where it has $H$-heads, region size of $R\times R$, and the average sparsity as $S$, which is the ratio for non-empty voxels\footnote{Our calculation is approximate because we assume non-empty voxels uniformly scatter in the space.}.
\begin{align}
	\label{eq:conv}
	\Omega(\text{Conv}) &= hwk^2C^2, \\
	\label{eq:sra}
	\Omega(\text{SRA}) &= 4ShwC^2 + 2HS^2R^2hwC,
\end{align}
As shown in the equations, the computational complexities for convolutions and SRA operations are all $O(hw)$, thus are both linear to the scale of input. However, the SRA operations have the linear factor of $S$, which is generally small due to the sparsity of point clouds. According to our statistics, $S$ roughly equals to 0.09 on Waymo Open Dataset with our voxelization. Such an analysis indicates that our SRA operations is efficient by properly exploiting the sparsity of LiDAR data.

\end{document}